\documentclass{article}
\usepackage{spconf,amsmath,graphicx,hyperref}

\usepackage{multirow}
\usepackage{tabularx}
\usepackage{booktabs}
\usepackage{amssymb}
\usepackage{makecell}
\usepackage{xcolor}
\usepackage{tikz}
\usepackage{pgfplots}
\pgfplotsset{compat=newest}
\usetikzlibrary{pgfplots.groupplots}

\title{SPARC: SuperPixel-Aware Region Contrastive Learning for Self-Supervised Dense Prediction}

\name{David Szczecina\sthanks{Indicates equal contribution, joint first-authorship.\\We acknowledge the support of the Natural Sciences and Engineering Research Council of Canada (NSERC) through the NSERC Discovery program and the NSERC CGRS-M program.},
Yuanpei Xiang\footnotemark[1],
Jitao Hu\footnotemark[1],
David Clausi,
Yuhao Chen,
Jason Deglint,
Paul Fieguth}

\address{Systems Design Engineering\\
    University of Waterloo\\
    Waterloo, Ontario, Canada}

\begin{document}

\maketitle

\begin{abstract}
Self-supervised learning (SSL) has become an effective approach for learning visual representations without manual annotations. Among SSL approaches, contrastive learning has been widely used for visual representation learning. However, existing contrastive SSL methods have focused primarily on image-level or pixel-level representation learning, while region-level representation learning remains less explored. We propose \textbf{SPARC}, a region-level contrastive learning framework that leverages superpixels to establish explicit correspondence between augmented image views. SPARC introduces a region contrastive branch that performs superpixel-based feature pooling and optimizes a region-level contrastive objective jointly with a global image-level objective. Under identical settings, SPARC consistently outperforms previous methods such as MoCo-v2 and DenseCL, achieving improvements of up to +9.79 mIoU for semantic segmentation and +4.88 AP for object detection. Ablation studies further demonstrate that region-level objectives produce the strongest performance. Thus, region-level contrastive learning is an effective approach for improving self-supervised visual pretraining for dense prediction tasks. Code repository can be accessed at \url{https://github.com/xRIPEIx/SPARC}.
\end{abstract}


\section{Introduction}
\label{sec:intro}

\begin{figure*}[ht]
    \centering
    \includegraphics[width=\linewidth]{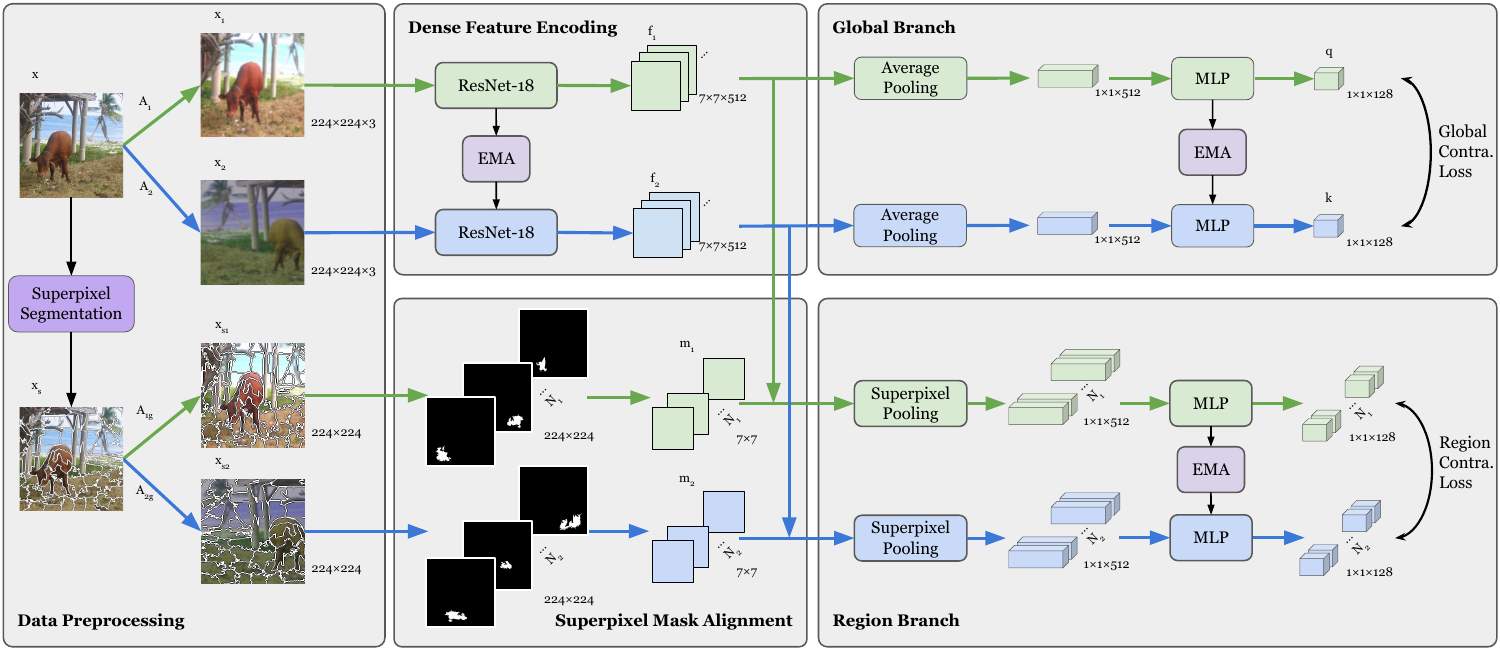}
    \caption{\textbf{SPARC Framework Overview.} SPARC's framework is inspired by DenseCL's double-branch design \cite{wang2021densecl}. We reframe it to adapt the region-level contrastive learning objective by introducing SLIC pre-segmentation, superpixel mask alignment, and superpixel pooling. Overall, SPARC performs feature and mask preparation through data preprocessing, dense feature encoding, and superpixel mask alignment, followed by contrastive learning branches. The global branch learns image-level representations, while the region branch uses superpixels to guide feature extraction and learn region-level representations.}
    \label{fig:SPARC_framework_overview}
\end{figure*}

Pre-training has become a fundamental paradigm in modern computer vision \cite{chavhan2023quality, hong2024training}, where models are first pre-trained on large-scale datasets and subsequently transferred to downstream tasks through fine-tuning \cite{he2022mae, li2023aligndet}. While supervised ImageNet pre-training has historically dominated this field \cite{krizhevsky2012imagnet}, recent advances in self-supervised learning (SSL) have demonstrated that visual representations can be learned directly from unlabelled data, achieving competitive or even superior performance on downstream benchmarks \cite{zhou2021ibot, oquab2024dinov2, szczecina2025pretrain}. 

Contrastive learning frameworks such as SimCLR \cite{chen2020simclr} and MoCo-v2~\cite{chen2020improved} have been particularly successful by learning image representations through instance discrimination, where different augmented views of the same image are encouraged to produce similar feature embeddings while remaining distinct from other images. These methods are designed and optimized for image-level representation learning \cite{wang2021densecl}. Thus, they are suited for image classification tasks, where a single semantic representation of an image is sufficient. 

However, for dense prediction tasks, including semantic segmentation and object detection, require substantially richer representations that preserve fine-grained spatial information  \cite{chaitanya2020contrastivelearninggloballocal, pinheiro2020unsupervisedlearningdensevisual}. The discrepancy between image-level pre-training objectives and pixel-level downstream prediction often limits the transferability of learned representations \cite{he2019rethinking, wang2021densecl}. Dense Contrastive Learning (DenseCL) addresses this limitation in \cite{wang2021densecl} by extending contrastive learning from global image-level representations to local feature representations. Instead of learning a single embedding for an image, DenseCL performs contrastive learning over spatial feature maps by identifying corresponding local features across augmented views. This narrows the gap between self-supervised pre-training and dense prediction tasks, producing stronger representations for semantic segmentation and object detection.Although DenseCL improves dense representation learning, its use of individual spatial features introduces a very fine-grained level of correspondence that is sensitive to local appearance variations and image noise \cite{xie2021pixpro}. This motivates exploring representations at an intermediate level of granularity, providing a natural middle ground for preserving local spatial structure while aggregating related features.

In this work, we investigate an intermediate representation between global image-level features and individual pixel features by introducing \textbf{SuperPixel-Aware Region Contrastive Learning (SPARC)}. Rather than contrasting individual spatial locations, SPARC constructs regions using SLIC \cite{achanta2012slic} superpixels and aggregates features within each region to form region-level representations. Correspondence across augmented views is established directly through transformed superpixel identities, providing explicit region-level positive pairs. By pooling features within corresponding regions before applying the contrastive objective, the framework preserves localized spatial information, and capture meaningful regional structure, while avoiding the reliance on feature-based pixel matching used by DenseCL \cite{wang2021densecl}.

To evaluate the proposed approach, SPARC, we pre-train on unlabelled images and transfer the learned representations to segmentation and detection benchmarks. Ablation studies are implemented to examine the effects of the balance between image-level and region-level objectives. Experimental results demonstrate consistent improvements over MoCo-v2 and DenseCL under identical pre-training settings. The proposed method, SPARC, produces representations that transfer more effectively to dense prediction tasks, highlighting region-level correspondence as an effective alternative to both image-level and pixel-level contrastive objectives. 

\section{Proposed Method}
\subsection{Overview}
SPARC reframes the DenseCL framework by replacing pixel-level correspondence with explicit region-level correspondence. The overall framework consists of data preprocessing, dense feature encoding, superpixel mask alignment, and the two contrastive learning branches. Similar to DenseCL, the global contrastive branch that learns image-level representations alongside a local contrastive branch. However, instead of performing contrastive learning on pixel-level, SPARC aggregates features within superpixel regions before computing the contrastive objective. For the rest of this section, we use \textit{region} to refer to superpixel regions.

\subsection{Feature and Mask Preparation}
\textbf{Data Preprocessing.} Instead of pre-segmenting all the images into fixed-size grid patches, we employ superpixels due to their nature of grouping similar neighbours together \cite{achanta2012slic}. As a result, we can maintain a more dynamic local relationship rather than relying on fixed patches \cite{lew2025suit}. For each training image, we first generate its superpixel segmentation map using the SLIC superpixel generation algorithm \cite{achanta2012slic}. 

The generated superpixel segmentation map, $X_s$, contains the region identities of the original RGB image, $X$. The original RGB image is then transformed by two different data augmentations, $A_1$ and $A_2$, to produce two augmented views, $X_1$ and $X_2$. In parallel, only the geometric augmentations of $A_1$ and $A_2$, $A_{1g}$ and $A_{2g}$, are applied to the corresponding superpixel segmentation map. This ensures that the regions remain spatially aligned with the augmented RGB images while preserving the consistent region identities across $X_{s1}$ and $X_{s2}$.

\textbf{Dense Feature Enconding.} The augmented RGB images, $X_1$ and $X_2$, are processed independently by the query and key backbone networks to produce dense feature maps. Similarly to the MoCo-v2 formulation \cite{chen2020improved}, the key backbone network is updated from the query backbone network using the exponential moving average (EMA). We then retain the spatial output of the fourth layer of ResNet-18, giving dense feature maps $f_1, f_2 \in \mathbb{R}^{7 \times 7 \times 512}$.

\textbf{Superpixel Mask Alignment.} The two transformed superpixel maps, $X_{s1}$ and $X_{s2}$, are decomposed into stacks of $N_{1}$ and $N_{2}$ binary masks, respectively. There is one binary mask per visible region, and $N_{1}$ and $N_{2}$ denote the number of visible regions in $X_{s1}$ and $X_{s2}$, respectively. Each binary mask marks the pixels belonging to a single original SLIC region ID. Then, all the binary masks are downsampled to $7 \times 7$ by using average pooling. This produces two stacks of soft region masks, $m_1 \in \mathbb{R}^{N_{1} \times 7 \times 7}$, and $m_2 \in \mathbb{R}^{N_{2} \times 7 \times 7}$. THis allows the spatial resolution of the superpixel masks matches the dense feature maps from the backbone network.

\subsection{Global Branch}
The global branch follows the DenseCL \cite{wang2021densecl} and MoCo-v2 \cite{chen2020improved} formulation without modification, as it already provides a strong image-level representation learning. The dense feature maps, $f_1$ and $f_2$, are downsampled to one pair of $512$-dimensional feature vectors using average pooling. The pooled features are then projected into $128$-dimensional image-level embeddings using a multilayer perceptron (MLP). Similarly to backbone networks, the key-side global projection head is updated from the query-side projection head using EMA. The positive pair is the two image-level feature embeddings from the same image, while the negative pairs are those from different images. The resulting image-level embeddings are optimized using the standard InfoNCE \cite{oord2018representation} contrastive loss to encourage consistency between different augmentations of the same image while separating representations from other images. The loss is defined as:
\begin{equation}
L_g = -\log \frac{ \exp\left(q \cdot k_{+} / \tau_g\right) } {\exp\left(q \cdot k_{+} / \tau_g\right)+ \sum_{k_{-} \in \beta}    \exp\left(q \cdot k_{-} / \tau_g\right)}
\end{equation}

\subsection{Region Branch}
The region branch learns representation at the level of individual local regions instead of treating each image as a single unit, multiple objects, or pixels. Given the dense feature maps, $f_1$, $f_2$, and the corresponding soft region masks $m_1$, $m_2$, we use each region's mask to weight and average the dense features, thus computing one $512$-dimensional feature vector per region. We named this as superpixel pooling, and the equation is defined as:
\begin{equation}
r_{v,n} = \frac{\sum_{i,j} m_{v,n}(i,j)\, f_v(i,j)}{\sum_{i,j} m_{v,n}(i,j)},
\qquad v \in \{1,2\},
\end{equation}
where $n \in \{1,\dots,N_v\}$ indexes the soft region masks of view $v$ and $(i,j)$ indexes spatial locations on the $7$ by the $7$ grid.

After superpixel pooling, each pooled region feature is subsequently projected from a $128$-dimensional embedding using a MLP. Similarly to the global branch, the key-side region projection head is updated from the query-sided projection head using EMA. Our method uses the superpixel identifier as ground truth region identity instead of defining correspondence explicitly. Positive pairs are defined as region embeddings originating from the same image and sharing the same transformed superpixel identifier across the two augmented views. All remaining region embeddings are treated as negative samples. The resulting region-level contrastive objective encourages semantically corresponding regions to produce similar feature representations while separating unrelated regions. The region contrastive loss is computed symmetrically in both directions between the two augmented views. The final region contrastive loss is obtained by averaging the two directions. The losses are defined as:
\begin{equation}
\resizebox{0.91\columnwidth}{!}{$\displaystyle
L_r^{q \to k}
=
-\frac{1}{\left|R_{q \to k}\right|}
\sum_{s \in R_{q \to k}}
\log
\frac{
    \exp\left((q_r^s)^\top k_{r,+}^s / \tau_r\right)
}{
    \exp\left((q_r^s)^\top k_{r,+}^s / \tau_r\right)
    +
    \sum_{k_r^- \in \beta_k^s}
    \exp\left((q_r^s)^\top k_r^- / \tau_r\right)
},$}
\end{equation}
\begin{equation}
\resizebox{0.91\columnwidth}{!}{$\displaystyle
L_r^{k \to q}
=
-\frac{1}{\left|R_{k \to q}\right|}
\sum_{s \in R_{k \to q}}
\log
\frac{
    \exp\left((k_r^s)^\top q_{r,+}^s / \tau_r\right)
}{
    \exp\left((k_r^s)^\top q_{r,+}^s / \tau_r\right)
    +
    \sum_{q_r^- \in \beta_q^s}
    \exp\left((k_r^s)^\top q_r^- / \tau_r\right)
}.$}
\end{equation}
\begin{equation}
L_r=\frac{1}{2}\left(L_r^{q \to k}+L_r^{k \to q}\right).
\end{equation}

\subsection{Training Objective}
Similar to DenseCL \cite{wang2021densecl}, the training objective combines the global image-level and region-level contrastive loss,
\begin{equation}
L = (1-\lambda)L_{\text{global}} + \lambda L_{\text{region}},
\end{equation}
where $\lambda$ controls the relative contribution of each objective. This joint optimization allows SPARC to preserve strong global semantic representations while simultaneously learning spatially localized region representations that are better suited for downstream dense prediction tasks.

\section{Experiments and Result}

\subsection{Datasets and Implementation Details}
We use two large-scale datasets for the pretraining experiments: MS COCO (118K images) \cite{lin2015microsoftcococommonobjects} and ImageNet100 (135k images) \cite{russakovsky2015imagenetlargescalevisual}. Downstream transfer is evaluated on PASCAL VOC \cite{PASCAL_VOC} for both segmentation and detection. For implementation, we use ResNet-18 as backbones. Following MoCo-v2 and DenseCL setup, models are pre-trained for 100 epochs using Adam ($lr=3 \times 10^{-4}$, weight decay $1 \times 10^{-4}$, batch size 64) with AMP enabled. The momentum encoder EMA follows a cosine schedule from 0.996 to 1.0. For DenseCL and SPARC, the local loss weight $\lambda$ is set to 0.5. For SPARC, we set the region contrastive temperature $\tau_r = 0.2$, pooling grid to $7 \times 7$, minimum region area to 1.0, and retain up to 16 regions per image. SLIC segmentation is configured with 100 segments, a compactness of 10, and $\sigma=1$.

\subsection{Downstream Evaluations and Results}
For semantic segmentation, we fine-tune an FCN on the VOC2012 train set and evaluate on the val set using AdamW (batch size 16, crop size 512, 520-pixel evaluation), reporting mIoU and pixel accuracy. For object detection, we fine-tune Faster R-CNN with FPN on VOC2012 using AdamW (batch size 2, $lr=1\times 10^{-4}$, weight decay $1\times 10^{-4}$, cosine decay) and report standard COCO-style metrics ($\text{AP}$, $\text{AP}_{50}$).

As shown in Table~\ref{tab:downstream_results_coco} and Table ~\ref{tab:downstream_results_IN100}, SPARC consistently outperforms both MoCo-v2 and DenseCL across all pre-training configurations and network backbones on PASCAL VOC. For semantic segmentation, SPARC yields substantial improvements, leading DenseCL by up to 5.32\% and MoCo-v2 by more than 9.0\% in mIoU. For object detection, SPARC outperforms MoCo-v2 and DenseCL, achieving gains exceeding 5.0\% in both AP and AP$_{50}$. These gains highlight the benefit of region-level correspondence, which captures semantically coherent structures and yields more transferable representations for dense prediction.

\begin{table}[t]
  \centering
  \footnotesize
  \renewcommand{\arraystretch}{1.13}
  \caption{Transfer performance on PASCAL VOC (in \%). Pre-trained on COCO with ResNet-18. SPARC consistently outperforms MoCo-v2 and DenseCL across all configurations.}
  \vspace{5pt}
  \label{tab:downstream_results_coco}
  \resizebox{\columnwidth}{!}{%
  \begin{tabular}{c|c|cc|cc}
  \hline
  \multirow{2}{*}{\shortstack{\textbf{Training} \\ \textbf{Epochs}}} & \multirow{2}{*}{\textbf{Method}} & \multicolumn{2}{c|}{\textbf{Segmentation}} & \multicolumn{2}{c}{\textbf{Detection}} \\
   & & \textbf{mIoU} & \textbf{Acc} & \textbf{AP} & \textbf{AP$_{\mathbf{50}}$} \\
  \hline
  \multirow{3}{*}{\shortstack{Pre-train: 50 \\ Finetune: 10}}
   & MoCo-v2  & 15.74 \tiny{$\pm$0.09} & 78.12 \tiny{$\pm$0.18} & 20.04 \tiny{$\pm$0.34} &
  41.97 \tiny{$\pm$0.55} \\
   & DenseCL & 20.21 \tiny{$\pm$0.41} & 79.63 \tiny{$\pm$0.25} & 20.97 \tiny{$\pm$0.06} &
  42.83 \tiny{$\pm$0.23} \\
   & \textbf{SPARC}   & \textbf{24.13} \tiny{$\pm$0.56} & \textbf{80.78} \tiny{$\pm$0.28} &
  \textbf{24.00} \tiny{$\pm$0.17} & \textbf{47.18} \tiny{$\pm$0.29} \\
  \hline
  \multirow{3}{*}{\shortstack{Pre-train: 100\\ Finetune: 10}}
   & MoCo-v2  & 18.01 \tiny{$\pm$0.19} & 79.08 \tiny{$\pm$0.26} & 20.63 \tiny{$\pm$0.26} &
  43.09 \tiny{$\pm$0.43} \\
   & DenseCL & 21.92 \tiny{$\pm$0.75} & 80.42 \tiny{$\pm$0.28} & 21.59 \tiny{$\pm$0.16} &
  43.59 \tiny{$\pm$0.18} \\
   & \textbf{SPARC}   & \textbf{26.55} \tiny{$\pm$0.44} & \textbf{81.70} \tiny{$\pm$0.27} &
  \textbf{24.37} \tiny{$\pm$0.17} & \textbf{47.74} \tiny{$\pm$0.44} \\
  \hline
   \multirow{3}{*}{\shortstack{Pre-train: 100\\ Finetune: 25}}
   & MoCo-v2  & 25.81 \tiny{$\pm$0.77} & 80.56 \tiny{$\pm$0.51} & 24.84 \tiny{$\pm$0.24} &
  48.99 \tiny{$\pm$0.30} \\
   & DenseCL & 29.57 \tiny{$\pm$0.53} & 81.64 \tiny{$\pm$0.92} & 25.48 \tiny{$\pm$0.12} & 48.90 \tiny{$\pm$0.28} \\
   & \textbf{SPARC}   & \textbf{33.78} \tiny{$\pm$0.16} & \textbf{83.23} \tiny{$\pm$0.21} &
  \textbf{27.67} \tiny{$\pm$0.30} & \textbf{51.77} \tiny{$\pm$0.36} \\
  \end{tabular}
  }
\end{table}
\setlength{\floatsep}{3pt}

\begin{table}[t]
  \centering
  \footnotesize
  \renewcommand{\arraystretch}{1.13}
  \caption{Same as Table \ref{tab:downstream_results_coco}, but pretrained on ImageNet100.}
  \vspace{5pt}
  \label{tab:downstream_results_IN100}
  \resizebox{\columnwidth}{!}{%
  \begin{tabular}{c|c|cc|cc}
  \hline
  \multirow{2}{*}{\shortstack{\textbf{Training} \\ \textbf{Epochs}}} & \multirow{2}{*}{\textbf{Method}} & \multicolumn{2}{c|}{\textbf{Segmentation}} & \multicolumn{2}{c}{\textbf{Detection}} \\
   & & \textbf{mIoU} & \textbf{Acc} & \textbf{AP} & \textbf{AP$_{\mathbf{50}}$} \\
  \hline
  \multirow{3}{*}{\shortstack{Pre-train: 50\\ Finetune: 10}}
   & MoCo-v2  & 13.68 \tiny{$\pm$0.47} & 76.74 \tiny{$\pm$0.22} & 18.73 \tiny{$\pm$0.37} & 39.97 \tiny{$\pm$0.54} \\
   & DenseCL & 17.70 \tiny{$\pm$0.63} & 78.24 \tiny{$\pm$0.35} & 20.20 \tiny{$\pm$0.20} & 41.53 \tiny{$\pm$0.23} \\
   & \textbf{SPARC}   & \textbf{21.48} \tiny{$\pm$0.35} & \textbf{79.86} \tiny{$\pm$0.31} & \textbf{23.38} \tiny{$\pm$0.38} & \textbf{46.06}
  \tiny{$\pm$0.64} \\
  \hline
  \multirow{3}{*}{\shortstack{Pre-train: 100\\ Finetune: 10}}
   & MoCo-v2  & 15.00 \tiny{$\pm$0.61} & 77.56 \tiny{$\pm$0.14} & 19.15 \tiny{$\pm$0.25} &
  40.66 \tiny{$\pm$0.29} \\
   & DenseCL & 18.61 \tiny{$\pm$0.53} & 78.80 \tiny{$\pm$0.18} & 20.41 \tiny{$\pm$0.19} &
  41.80 \tiny{$\pm$0.35} \\
   & \textbf{SPARC}   & \textbf{22.68} \tiny{$\pm$0.39} & \textbf{80.49} \tiny{$\pm$0.30} &
  \textbf{24.03} \tiny{$\pm$0.18} & \textbf{46.98} \tiny{$\pm$0.22} \\
  \hline
  \multirow{3}{*}{\shortstack{Pre-train: 100\\ Finetune: 25}}
   & MoCo-v2  & 21.14 \tiny{$\pm$0.27} & 79.49 \tiny{$\pm$0.21} & 23.60 \tiny{$\pm$0.11} &
  47.25 \tiny{$\pm$0.18} \\
   & DenseCL & 25.61 \tiny{$\pm$0.34} & 80.92 \tiny{$\pm$0.23} & 24.51 \tiny{$\pm$0.18} &
  47.54 \tiny{$\pm$0.25} \\
   & \textbf{SPARC}   & \textbf{30.93} \tiny{$\pm$0.23} & \textbf{82.35} \tiny{$\pm$0.19} &
  \textbf{27.31} \tiny{$\pm$0.09} & \textbf{51.09} \tiny{$\pm$0.20} \\
  \end{tabular}
  }
\end{table}

\subsection{Effect of the Region Loss Coefficient \texorpdfstring{$\lambda$}{lambda}}
The parameter $\lambda$ controls the relative contribution of the image-level contrastive loss and the proposed region-level contrastive loss within the overall training objective. The motivation for this ablation study is to understand the necessity of incorporating both image-level and region-level contrastive learning objectives to optimize the feature representation learning. We evaluated the model performance across a range of $\lambda$ values from 0 to 1.0.

\begin{figure}[t]
\centering
\providecolor{series1}{HTML}{2A78D6}
\providecolor{series2}{HTML}{EB6834}
\providecolor{series3}{HTML}{2CA02C}
\providecolor{ink2}{HTML}{52514E}
\providecolor{grid}{HTML}{E1E0D9}
\providecolor{rule}{HTML}{C3C2B7}
\begin{tikzpicture}
\begin{groupplot}[
  group style={group size=2 by 1, horizontal sep=1.3cm, vertical sep=0.75cm, xticklabels at=edge bottom, xlabels at=edge bottom},
  width=4.5cm, height=4.5cm,
  xlabel={loss weight $\lambda$}, xlabel style={yshift=2pt},
  xmin=-0.04, xmax=1.04,
  xtick={0,0.1,0.3,0.5,0.7,0.9,1.0},
  xticklabels={0,0.1,0.3,0.5,0.7,0.9,1},
  axis line style={rule, thin},
  axis x line*=bottom, axis y line*=left,
  tick style={rule, thin}, tick label style={font=\footnotesize, color=ink2},
  label style={font=\footnotesize, color=ink2},
  ymajorgrids, grid style={grid, thin},
  clip=false,
  legend style={font=\footnotesize, draw=none, fill=none, at={(0.03,0.05)}, anchor=south west},
  legend cell align=left,
  error bars/y dir=both, error bars/y explicit,
  error bars/error bar style={thin},
  every axis plot/.append style={line width=1.4pt, mark size=2.2pt},
  xtick={0,0.3,0.5,0.7,1.0}, xticklabels={0,0.3,0.5,0.7,1},
  tick label style={font=\scriptsize, color=ink2},
]
\nextgroupplot[ylabel={VOC2012 mIoU (\%)}, ymin=26, ymax=41, ytick={26,28,...,40}, legend columns=-1, legend to name=lambdalegend, legend style={draw=none, fill=none, /tikz/every even column/.append style={column sep=10pt}}]
\addplot[series1, mark=*, mark options={fill=series1}]
  coordinates {
    (0,34.41)+-(0,0.31) (0.1,36.75)+-(0,0.22) (0.3,38.41)+-(0,0.06)
    (0.5,38.80)+-(0,0.39) (0.7,38.88)+-(0,0.45) (0.9,38.78)+-(0,0.35)
    (1.0,38.32)+-(0,0.29)
  };
\addlegendentry{SPARC}
\addplot[series2, mark=square*, mark options={fill=series2}]
  coordinates {
    (0,34.19)+-(0,0.32) (0.1,36.49)+-(0,0.43) (0.3,37.50)+-(0,0.50)
    (0.5,37.38)+-(0,0.19) (0.7,36.61)+-(0,0.39) (0.9,33.66)+-(0,0.68)
    (1.0,27.56)+-(0,1.47)
  };
\addlegendentry{DenseCL}
\addplot[series3, only marks, mark=triangle*, mark size=3.5pt, mark options={fill=series3}]
  coordinates {(0,34.24)+-(0,0.52)};
\addlegendentry{MoCo-v2}
\nextgroupplot[ylabel={VOC2012 AP (\%)}, ymin=16, ymax=27, ytick={16,18,...,26}]
\addplot[series1, mark=*, mark options={fill=series1}]
  coordinates {
    (0,23.66)+-(0,0.26) (0.1,24.47)+-(0,0.29) (0.3,25.10)+-(0,0.21)
    (0.5,25.41)+-(0,0.16) (0.7,25.53)+-(0,0.11) (0.9,25.30)+-(0,0.14)
    (1.0,25.55)+-(0,0.15)
  };
\addplot[series2, mark=square*, mark options={fill=series2}]
  coordinates {
    (0,23.66)+-(0,0.32) (0.1,24.04)+-(0,0.16) (0.3,24.21)+-(0,0.24)
    (0.5,24.03)+-(0,0.18) (0.7,23.52)+-(0,0.22) (0.9,21.74)+-(0,0.19)
    (1.0,18.36)+-(0,0.89)
  };
\addplot[series3, only marks, mark=triangle*, mark size=3.5pt, mark options={fill=series3}]
  coordinates {(0,23.68)+-(0,0.14)};
\end{groupplot}
\end{tikzpicture}\\[2pt]
\ref{lambdalegend}
\caption{\textbf{Semantic Segmentation and Object Detection Performance} between SPARC and DenseCL across different $\lambda$ values. At $\lambda=0$, both methods use only the image-level contrastive objective, corresponding to the MoCo-v2 baseline. At $\lambda=1$, only the region-level or pixel-level contrastive objective is used for the proposed SPARC and DenseCL, respectively. The proposed SPARC's pure region-level contrastive learning significantly outperforms pure pixel-level and image-level contrastive learning frameworks.}
\label{fig:lambda-all}
\end{figure}
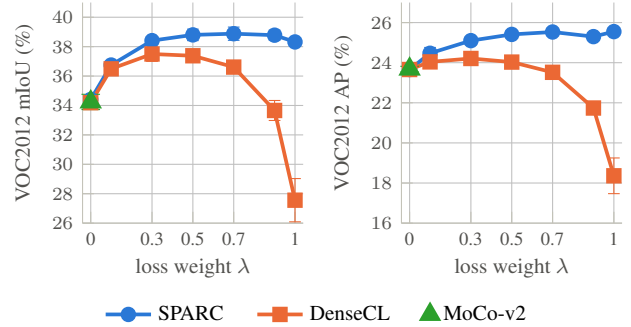

As shown in Figure \ref{fig:lambda-all}, the proposed method SPARC outperforms DenseCL across all different $\lambda$ values and MoCo-v2. It also illustrates that pure region-level significantly outperforms pure pixel-level, and it is better than pure image-level. Based on this ablation study result, we conclude that implementing region-level contrastive learning framework can lead to better downstream application performance.

\section{Conclusion}
In this work, we presented SPARC, a region-level self-supervised learning framework for visual representation learning that extends DenseCL by replacing pixel-level correspondence with explicit superpixel-based region correspondence. By performing contrastive learning on pooled region representations while jointly optimizing a global image-level objective, SPARC learns feature representations that are better aligned with downstream dense prediction tasks.

Experimental evaluation on semantic segmentation and object detection benchmarks demonstrates consistent improvements over DenseCL under identical training settings. Ablation studies further show that region-level objectives provides the strongest performance. Overall, these results suggest that region-level contrastive learning provides an effective and computationally efficient direction for improving self-supervised visual pre-training for dense prediction tasks. Future work will investigate replacing SLIC superpixels with learned region generators, extending SPARC to modern transformer-based self-supervised learning frameworks, and evaluating the approach on larger-scale pre-training datasets, and additional dense prediction tasks.

\vfill\pagebreak

\bibliographystyle{IEEEbib}
\bibliography{main}

\end{document}